\documentclass[runningheads]{llncs}
\usepackage{graphicx}

\usepackage{hyperref}
\usepackage{adjustbox}
\usepackage{amssymb}
\usepackage{xcolor-material}
\usepackage{tikz}
\usepackage{xcolor}
\usepackage{mathtools}
\usepackage{ragged2e}
\usepackage[caption=false]{subfig}
\usepackage{color}
\usepackage{floatrow}
\begin{document}
\title{Spatial Temporal Transformer Network for Skeleton-based Action Recognition}
\titlerunning{ST-TR Network for Skeleton-based Action Recognition}
%
\author{Chiara Plizzari \and
Marco Cannici \and
Matteo Matteucci}
\authorrunning{C. Plizzari et al.}
%
\institute{
Politecnico di Milano, Milano, Italy \\ \email{chiara.plizzari@mail.polimi.it}\\ \email{\{marco.cannici, matteo.matteucci\}@polimi.it}}
\maketitle              
\begin{abstract}
Skeleton-based human action recognition has achieved a great interest in recent years, as skeleton data has been demonstrated to be robust to illumination changes, body scales, dynamic camera views, and complex background. 
Nevertheless, an effective encoding of the latent information underlying the 3D skeleton is still an open problem. 
In this work, we propose a novel Spatial-Temporal Transformer network (ST-TR) which models dependencies between joints using the Transformer \textit{self-attention} operator. In our ST-TR model, a Spatial Self-Attention module (SSA) is used to understand intra-frame interactions between different body parts, and a Temporal Self-Attention module (TSA) to model inter-frame correlations. The two are combined in a two-stream network which outperforms state-of-the-art models using the same input data on both NTU-RGB+D 60 and NTU-RGB+D 120.

\end{abstract}

\section{Introduction}

Skeleton-based activity recognition is achieving increasing interest in recent years thanks to advances in 3D skeleton pose estimation devices, both in terms of accuracy and resolution. Algorithms and neural architectures for extracting context-aware fine-grained spatial-temporal features, capable of unlocking the true potential of skeleton based action recognition, however, are still lacking in the literature. 
The most widespread method to perform skeleton-based action recognition has become Spatial-Temporal Graph Convolutional Network (ST-GCN)~\cite{yan2018spatial}, since, being an efficient representation of non-Euclidean data, it is able to effectively capture spatial (intra-frame) and temporal (inter-frame) information. 
However, ST-GCN models have some structural limitations, some of them already addressed in~\cite{dirgraph,Shi2018TwoStreamAG,shift,disent}: (i) The topology of the graph representing the human body is fixed for all layers and all the actions, preventing the extraction of rich representations.
(ii) Both the spatial and temporal convolutions are implemented from a standard 2D convolution. As such, they are limited to operate in a local neighborhood. 
(iii) As a consequence of (i) and (ii), correlations between body joints not linked in the human skeleton, e.g., the left and right hands, are underestimated even if relevant in actions such as ``clapping".

In this paper, we face all these limitations by employing a modified Transformer self-attention operator. Despite being originally designed for Natural Language Processing (NLP) tasks, the flexibility of the Transformer self-attention~\cite{attention} in modeling long-range dependencies, make this model a perfect solution to tackle ST-GCN weaknesses. Recently, Bello et al. in~\cite{DBLP:journals/corr/abs-1904-09925} employed self-attention on image pixels to overcome the locality of the convolution operator. 
In our work, we aim to apply the same mechanism to joints representing the human skeleton, with the goal of extracting adaptive low-level features modeling interactions in human actions both in space, through a Spatial Self-Attention module (SSA), and time, through a Temporal Self-Attention module (TSA) module. Authors of~\cite{san} also proposed a Self-Attention Network (SAN) to extract long-term semantic information; however, since they focus on temporally segmented clips, they solve the locality limitations of convolution only partially. 

Contributions of this paper are summarized as follows: 
\begin{itemize}
    \item[\textbullet] We propose a novel two-stream Transformer-based model, employing \textit{self-attention} on both the spatial and temporal dimensions
    \item[\textbullet] We design a \textit{Spatial Self-Attention} (SSA) module to dynamically build links between skeleton joints, representing the relationships between human body parts, conditionally on the action and independently from the natural human body structure. On the temporal dimension, we introduce a  \textit{Temporal Self-Attention} (TSA) module to study the dynamics of joints along time too~\footnote{Code at {\url{{https://github.com/Chiaraplizz/ST-TR}}}}
    \item[\textbullet] Our model outperforms the ST-GCN \cite{yan2018spatial} baseline, and outperforms previous state-of-the-art methods using the same input data on NTU-RGB+D \cite{ntu,ntu120}.
\end{itemize}

\begin{figure*}[t]
\centering
\subfloat[Spatial Self-Attention \label{2a}]{\begin{minipage}[b]{.51\linewidth}
    \includegraphics[width=\textwidth]{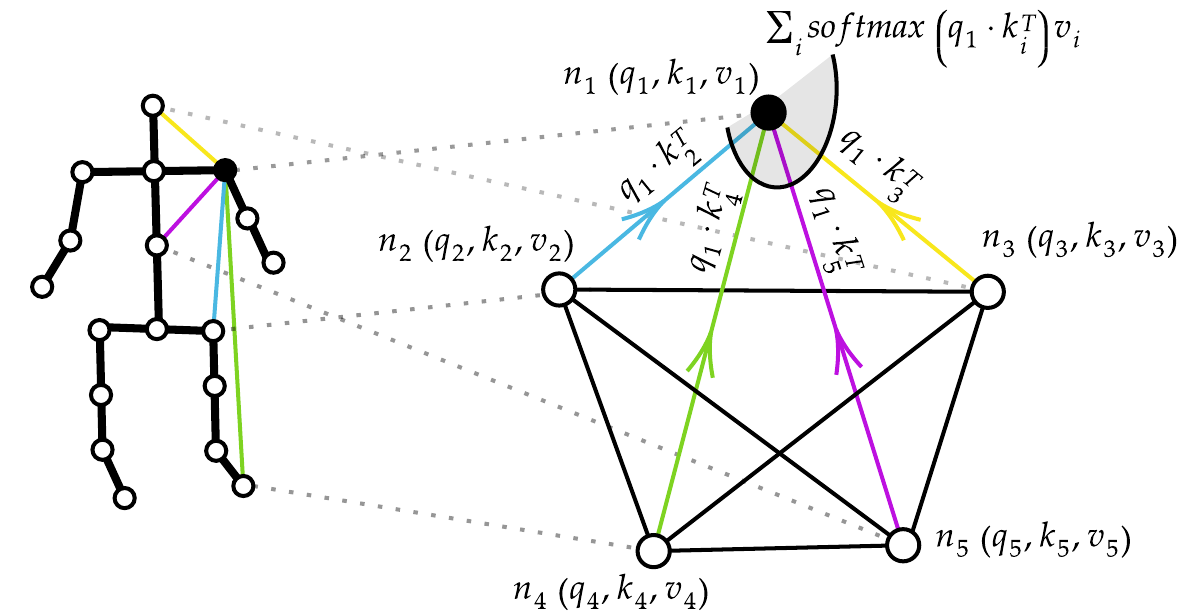}
\end{minipage}}
\subfloat[Temporal Self-Attention \label{2b}]{\begin{minipage}[b]{.5\linewidth}
    \includegraphics[width=\textwidth]{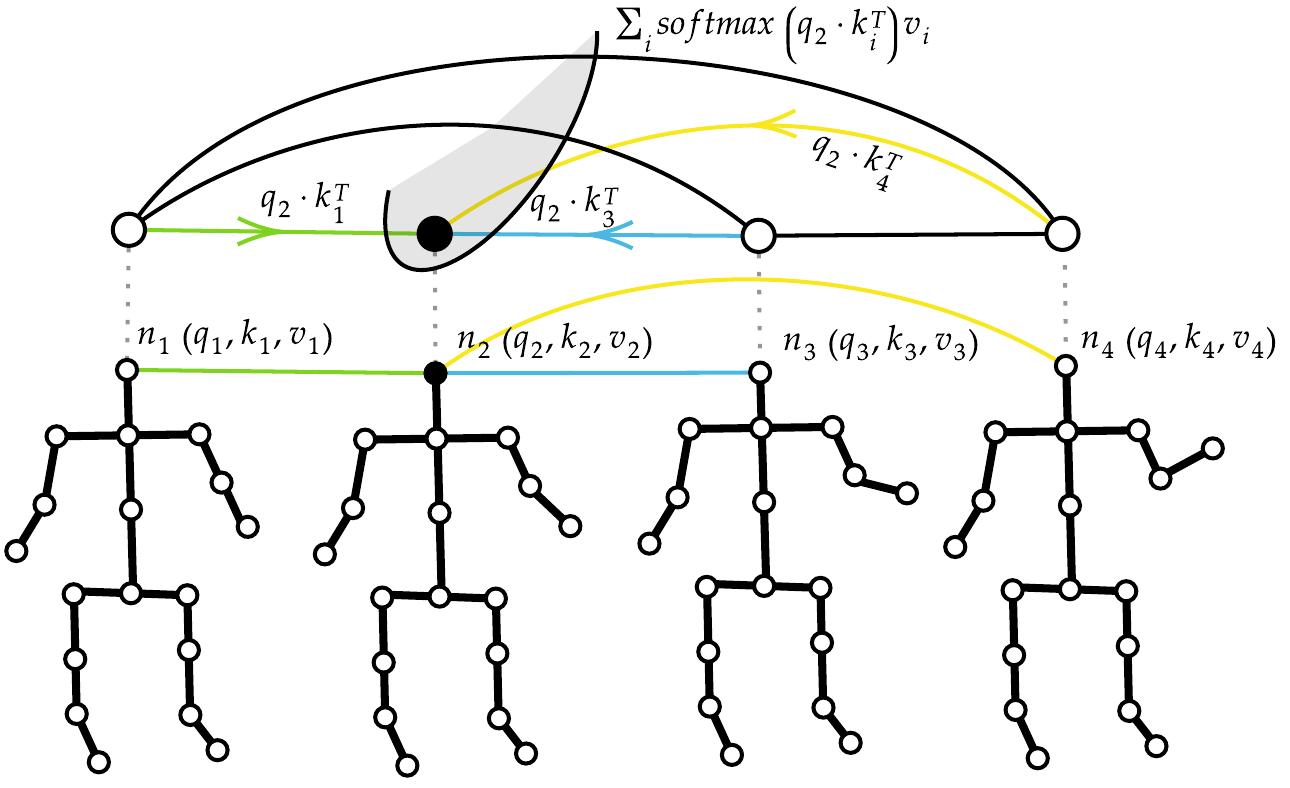}
\end{minipage}}
\caption{{Spatial Self-Attention (SSA)} and {Temporal Self-Attention (TSA)}. Self-attention operates on each pair of nodes by computing a weight for each of them representing the strength of their correlation. Those weights are used to score the contribution of each body joint $n_i$, proportionally to how relevant the node is w.r.t. the other ones. }
\label{SSA_TSA}
\end{figure*}

\section{Spatial Temporal Transformer Network}

We propose \textit{Spatial Temporal Transformer (ST-TR)}, an architecture using the Transformer self-attention mechanism to operate both on space and time. We develop two modules, \textit{Spatial Self-Attention (SSA)} and \textit{Temporal Self-Attention (TSA)}, each one focusing on extracting correlations in one of the two dimensions.

\subsection{Spatial Self-Attention (SSA)}

The SSA module applies self-attention \textit{inside each frame} to extract low-level features embedding the relations between body parts, i.e., computing correlations between each pair of joints in every single frame independently, as depicted in Figure~\ref{SSA_TSA}a. 
Given the frame at time $t$, for each node $i^t$ of the skeleton, a \textit{query} vector  $\mathbf{q}^t_i \in \mathbb{R}^{dq}$, a \textit{key} vector $\mathbf{k}^t_i \in \mathbb{R}^{dk}$ and a \textit{value} vector $\mathbf{v}^t_i \in \mathbb{R}^{dv}$ are first computed by applying trainable linear transformations to the node features $\mathbf{n}_i^t \in \mathbb{R}^{C_{in}}$, shared across all nodes, of parameters $\mathbf{W}_q \in \mathbb{R}^{C_{in} \times dq}$, $\mathbf{W}_k \in \mathbb{R}^{C_{in} \times dk}$, $\mathbf{W}_v \in \mathbb{R}^{C_{in} \times dv}$. Then, for each pair of body nodes $(i^t, j^t)$, a \textit{query-key dot product} is applied to obtain a weight $\alpha_{ij}^t \in \mathbb{R}$ representing the strength of the correlations between the two nodes. The resulting score $\alpha^t_{ij}$ is used to weight each joint value $\textbf{v}^t_j$, and a weighted sum is computed to obtain a new embedding for node $i^t$, as in the following:
\begin{equation}\label{eq5}
     \alpha^t_{ij}={\mathbf{q}}^t_{i} \cdot {{\mathbf{k}}^t_{j}}^T, \forall t \in T  ,\qquad
    {\mathbf{z}}^t_i=\sum_j{softmax_j \left(\frac{\alpha^t_{ij}}{\sqrt{d\textsubscript{k}}} \right)  {\mathbf{v}}^t_j}
\end{equation}
where ${\mathbf{z}^t_i} \in \mathbb{R}^{C_{out}}$ (with $C_{out}$ the number of output channels) constitutes the new embedding of node $i^t$.

Multi-head self-attention is finally applied by repeating this embedding extraction process $H$ times, each time with a different set of learnable parameters. The set $(\mathbf{z}^t_{i_1}, ..., \mathbf{z}^t_{i_H})$ of node embeddings thus obtained, all referring to the same node $i^t$, is finally combined with a learnable transformation, i.e., $concat(\mathbf{z}^t_{i_1},...,\mathbf{z}^t_{i_H})\cdot\mathbf{W}_o$, and constitutes the output features of SSA.

Thus, as shown in Figure~\ref{SSA_TSA}a, the relations between nodes (i.e., the $\alpha_{ij}^t$ scores) are dynamically \textit{predicted} in SSA; the correlation structure is not fixed for all the actions, but it changes adaptively for each sample. SSA operates similar to a graph convolution on a fully connected graph where, however, the kernel values (i.e., the $\alpha_{ij}^t$ scores) are predicted dynamically based on the skeleton pose.

\subsection{Temporal Self-Attention (TSA)}\label{sec:st}

Along the temporal dimension, the dynamics of each joint is studied separately \textit{along all the frames}, i.e., each single node is considered as independent and correlations between frames are computed by comparing features of the same body joint along the temporal dimension (see Figure~\ref{SSA_TSA}b). The formulation is symmetrical to the one reported in Equation \eqref{eq5} for SSA:
\begin{equation}
    \alpha^v_{ij} =\mathbf{q}^v_{i} \cdot {\mathbf{k}^v_{j}} \quad \forall v \in V ,\qquad
    \mathbf{z}^v_{i} =\sum_j{softmax_j\left(\frac{\alpha^v_{ij}}{\sqrt{d\textsubscript{k}}}\right)  \mathbf{v}^v_j}
\end{equation}
where $i^v, j^v$ indicate the same joint $v$ in two different instants $i, j$, $\alpha^v_{ij} \in \mathbb{R}$, $\mathbf{q}^v_i \in \mathbb{R}^{dq}$ is the query associated to $i^v$, $\mathbf{k}^v_j \in \mathbb{R}^{dk}$ and $\mathbf{v}^v_j \in \mathbb{R}^{dv}$ are the key and value associated to joint $j^v$ (all computed using trainable linear transformations as in SSA), and $\mathbf{z}^v_i \in \mathbb{R}^{C_{out}}$ is the resulting node embedding. An illustration of TSA is depicted in Figure~\ref{SSA_TSA}b. Multi-head attention is applied as in SSA.
The network, by extracting inter-frame relations between nodes in time, can learn to correlate frames apart from each other (e.g., nodes in the first frame with those in the last one), capturing discriminant features that are not otherwise possible to capture with a standard convolution, being this limited by the kernel size.

\subsection{Two-Stream Spatial Temporal Transformer Network}\label{2s}

\begin{figure}[t]
    \centering
    \includegraphics[width=0.75\textwidth]{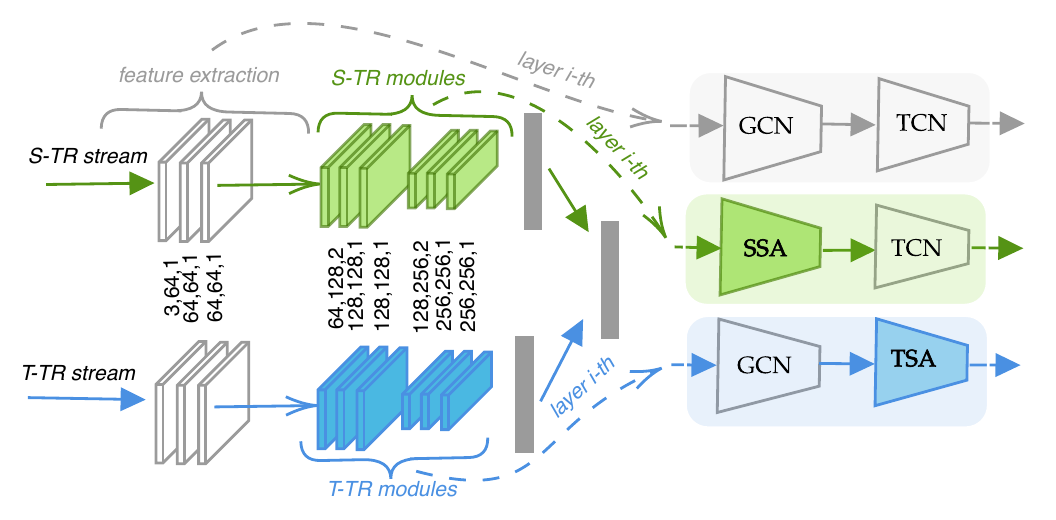}
    \caption{The 2s-ST-TR architecture. On each stream, the first three layers extract simple features. On the S-TR stream, at each subsequent layer, SSA is used to extract spatial information, followed by a 2D convolution on time dimension (TCN). On the T-TR stream, at each subsequent layer, TSA is used to extract temporal information, while spatial features are extracted by a standard graph convolution (GCN) \cite{yan2018spatial}}
    \label{architecture}
\end{figure}

To combine the SSA and TSA modules, a two-stream architecture named 2s-ST-TR is used, as similarly proposed by Shi et al. in~\cite{Shi2018TwoStreamAG} and~\cite{dirgraph}. In our formulation, the two streams differentiate on the way the self-attention mechanism is applied: SSA operates on the spatial stream (named S-TR stream), while TSA on the temporal one (named T-TR stream). 
On both streams, simple features are first extracted through a three-layers residual network, where each layer processes the input on the spatial dimension through graph convolution (GCN), and on the temporal dimension through a standard 2D convolution (TCN), as in ST-GCN~\cite{yan2018spatial}. SSA and TSA are applied on the S-TR and on the T-TR stream in substitution to the GCN and TCN feature extraction modules respectively (Figure~\ref{architecture}). Each stream is trained using the standard cross-entropy loss, and the sub-networks outputs are eventually fused together by summing up their softmax output scores to obtain the final prediction, as in~\cite{Shi2018TwoStreamAG,dirgraph}. 

\paragraph*{\textbf{Spatial Transformer Stream (S-TR).}} In the spatial stream (Figure \ref{architecture}), self-attention is applied at the skeleton level through a SSA module, which focuses on spatial relations between joints, and then its output is passed to a 2D convolutional module with kernel $K_t$ on temporal dimension (TCN), as in \cite{yan2018spatial}, to extract temporally relevant features, i.e., $\textbf{S-TR}(x)= Conv_{2D(1\times{K_t})}(\textbf{{SSA}}(x))$.
Following the original Transformer structure, the input is pre-normalized passing through a Batch Normalization layer~\cite{ioffe2015batch,bn_tr}, and skip connections are used, which sum the input to the output of the SSA module.

\paragraph*{\textbf{Temporal Transformer Stream (T-TR).}} 
The temporal stream, instead, focuses on discovering inter-frame temporal relations. Similarly to the S-TR stream, inside each T-TR layer, a standard graph convolution sub-module \cite{yan2018spatial} is followed by the proposed Temporal Self-Attention module, i.e., $\textbf{T-TR}(x)= \textbf{TSA}(GCN(x))$. In this case, TSA operates on graphs linking the same joint along all the time dimension (e.g., all left feet, or all right hands).

\section{Model Evaluation}
\subsection{Datasets}

\paragraph*{\textbf{\textit{NTU RGB+D 60 and NTU RGB+D 120}.}}{The NTU RGB+D 60 (NTU-60) dataset is a large-scale benchmark for 3D human action recognition \cite{ntu}.  
Skeleton information consists of 3D coordinates of $25$ body joints and a total of $60$ different action classes. The NTU-60 dataset follows two different criteria for evaluation. In \textit{Cross-View Evaluation} (X-View), the data is split according to the camera from which the action is taken, while in \textit{Cross-Subject Evaluation} (X-Sub) according to the subject performing the action. 
NTU-RGB+D 120~\cite{ntu120} (NTU-120) is an extension of NTU-60, with a total of $113,945$ videos and $120$ classes.
It follows two evaluation criteria: \textit{Cross-Subject Evaluation (X-Sub)} is the same used in NTU-60, while in \textit{Cross-Setup Evaluation (X-Set)} training and testing samples are split based on the parity of the camera setup IDs.  } 

\subsection{Experimental Settings} Using PyTorch framework, we trained our models for a total of $120$ epochs with batch size $32$ and SGD as optimizer. The learning rate is set to $0.1$ at the beginning and then reduced by a factor of $10$ at the epochs \{$60$, $90$\}.
Moreover, we preprocessed the data with the same procedure used by Shi et al. in~\cite{Shi2018TwoStreamAG} and~\cite{dirgraph}. In order to avoid overfitting, we also used \textit{DropAttention}~\cite{Lin2019DropAttentionAR}, a dropout technique introduced by Zehui et al.~\cite{Lin2019DropAttentionAR} for regularizing attention weights in Transformer.
In all of these experiments, the \textit{number of heads} for multi-head attention is set to $8$, and $d_q, d_k, d_v$ embedding dimensions to $0.25 \times C_{out}$ in each layer, as in \cite{DBLP:journals/corr/abs-1904-09925}. We did not perform a grid search on these parameters.

\subsection{{Results}}
To verify the effectiveness of our SSA and TSA modules, we compare separately the S-TR stream and T-TR stream against the ST-GCN \cite{yan2018spatial} baseline, whose results on NTU-60~\cite{ntu} are reported in Table \ref{table:1} using our learning rate scheduling.

\begin{table}[t]
\centering
\caption{a) Accuracy (\%) comparison of S-TR and T-TR, and their combination (ST-TR) on NTU-60, w and w/o bones. b) Parameters of SSA and TSA modules}
\label{table:1}
\begin{minipage}[t]{0.55\linewidth}{\small (a)}
\centering
\begin{adjustbox}{width=0.7\columnwidth, margin=0ex 1ex 0ex 0ex}
\subfloat{

    \begin{tabular}{lccc}
    
    \hline\noalign{\smallskip}
    \textbf{Method} & \textbf{Bones} & X-Sub & X-View \\\noalign{\smallskip}
    \hline
    \noalign{\smallskip}
        ST-GCN \cite{yan2018spatial} & & 85.7 & 92.7  \\
    \hline
    S-TR &  &86.4 & 94.0 \\
    T-TR  &   &86.0 & 93.6 \\
    \hline
    ST-TR  & &88.7 & 95.6 \\
    \hline
    
    S-TR   & \checkmark &87.9 & 94.9 \\
    T-TR  &\checkmark &87.3 & 94.1\\
    \hline
    ST-TR  & \checkmark& 89.9 & 96.1\\

    \hline
    \end{tabular}
}
\end{adjustbox}
\end{minipage}
\hfill
\begin{minipage}[t]{0.34\linewidth}{\small (b)}
\begin{adjustbox}{width=0.8\columnwidth, margin=0ex 1ex 0ex 0ex}
\centering
\subfloat{

    \begin{tabular}{lc}
    \hline\noalign{\smallskip}
    \textbf{Module} &Params [$\times 10^4$] \\\noalign{\smallskip}

    \hline\noalign{\smallskip}

    GCN~\cite{yan2018spatial} &19.9 \\
    SSA &17.8 
    \\
        \hline

    TCN~\cite{yan2018spatial} &59.0 \\
    TSA &17.7  \\
    \hline
    \end{tabular}
}
\end{adjustbox}
\label{table:1ab}
\end{minipage}
\end{table}
\setlength{\tabcolsep}{1.4pt}

As far as it concerns the SSA, S-TR outperforms the baseline by $0.7\%$ on X-Sub, and by $1.3 \%$ on X-View, demonstrating that self-attention can be used in place of graph convolution, increasing the network performance while also decreasing the number of parameters. On NTU-60 the S-TR stream achieves slightly better performance (+0.4\%) than the T-TR stream, on both X-View and X-Sub (Table \ref{table:1}a). This can be motivated by the fact that SSA in S-TR operates on 25 joints only, while on the temporal dimension the number of correlations is proportional to the huge number of frames. 
Table \ref{table:1}b shows the difference in terms of parameters between a single GCN (TCN) and the corresponding SSA (TSA) module, with $C_{in}=C_{out}=256$. 
Especially on the temporal dimension, TSA results in a decrease in parameters, introducing $41.3 \times 10^4$ less than TCN. The combination of the two streams achieves 88.7\% of accuracy on X-Sub and 95.6\% of accuracy on X-View, outperforming the baseline ST-GCN by up to $3\%$ and surpassing other two-stream architectures~(Table~\ref{table:3}). Classes that benefit the most from self-attention are “playing with phone”, “typing”, and “cross hands” on S-TR, and those involving long-range relations or two people, i.e., “hugging”, “point finger”, “pat on back”, on T-TR. These require to correlate along the entire action, giving empirical insight on the advantage of the proposed method.
 
As adding bones information demonstrated leading to better results in previous works \cite{dirgraph,Shi2018TwoStreamAG}, we also studied the effect of our self-attention module on combined joint and bones information. For each node $\mathbf{v}_1=(\mathbf{x}_1, \mathbf{y}_1, \mathbf{z}_1)$ and $\mathbf{v}_2=(\mathbf{x}_2,\mathbf{y}_2,\mathbf{z}_2)$, the bone connecting the two is calculated as 
    $\mathbf{b}_{\mathbf{v}_1,\mathbf{v}_2}=(\mathbf{x}_2-\mathbf{x}_1,\mathbf{y}_2-\mathbf{y}_1,\mathbf{z}_2-\mathbf{z}_1)$.
Joint and bone information are concatenated along the channel dimension and then fed to the network. At each layer, the size of the input and output channels is doubled as in \cite{dirgraph,Shi2018TwoStreamAG}. The performance results are shown again in Table~\ref{table:1}a; all previous configurations improve when bones are added as input. The latter fact highlights the flexibility of our method, which is capable of adapting to different input types and network configurations. 

\begin{table}[t]

\caption{Comparison with state-of-the-art accuracy (\%) of S-TR, T-TR, and their combination (ST-TR) on NTU-60 (a) and NTU-120 (b)}
\label{table:3}

\begin{minipage}[t]{0.51\linewidth}{\small (a)} \centering
\begin{adjustbox}{width=0.88\columnwidth, margin=0ex 1ex 0ex 0ex}
\subfloat{

    \begin{tabular}{lccc}
    \hline\noalign{\smallskip}
\multicolumn{4}{c}{\textbf{NTU-60}}\\
\cline{1-4}\noalign{\smallskip}
\textbf{Method} & \textbf{Bones} & X-Sub & X-View\\    \noalign{\smallskip}
    \hline
    \noalign{\smallskip}
    ST-GCN \cite{yan2018spatial} & & 81.5 & 88.3\\
     1s-AGCN \cite{Shi2018TwoStreamAG}\cite{disent} & & 86.0 & 93.7 \\
     
    1s Shift-GCN \cite{shift}  & &87.8 & 95.1 \\
     SAN \cite{san} &&87.2 & 92.7 \\
    \hline
    ST-TR (Ours)  & &\textbf{88.7} & \textbf{95.6} \\
    
\hline\noalign{\smallskip}
    2s-AGCN \cite{Shi2018TwoStreamAG}  & \checkmark& 88.5 & 95.1 \\
    DGCNN \cite{dirgraph}  &\checkmark &  89.9 & 96.1 \\
    2s Shift-GCN \cite{shift}   & \checkmark &89.7 & 96.0\\
MS-G3D \cite{disent}  &\checkmark& \textbf{91.5} & \textbf{96.2} \\

\hline
    ST-TR (Ours) & \checkmark & {{89.9}} & {{96.1}}
    \\
    \hline
    \end{tabular}
    }
    \end{adjustbox}
\end{minipage}
\hfill
\begin{minipage}[t]{0.47\linewidth}{\small (b)}
\begin{adjustbox}{width=0.9\columnwidth, margin=0ex 1ex 0ex 0ex}
\centering
\subfloat{

    \begin{tabular}{lcc}
    \hline\noalign{\smallskip}
    \multicolumn{3}{c}{\textbf{NTU-120}}\\
    \cline{1-3}\noalign{\smallskip}
    {\textbf{Method}} & X-Sub & X-Set\\
    \noalign{\smallskip}
    \hline\noalign{\smallskip}
    {ST-LSTM \cite{st-lstm}} & 55.7 & 57.9 \\
    {GCA-LSTM \cite{gca}} & 61.2 & 63.3 \\
    {RotClips+MTCNN \cite{lcr}} &62.2&61.8\\

    {Pose Evol. Map \cite{bpe}} & 64.6 & 66.9\\
    1s Shift-GCN \cite{shift} & 80.9 & 83.2 \\
    \hline
    {S-TR} (Ours) & 78.6 & 80.7 \\
    {T-TR} (Ours) &   78.4 & 80.5 \\

    \hline
    {ST-TR} (Ours)  & \textbf{81.9} & \textbf{84.1} \\
    \hline
    \end{tabular}
    }
\end{adjustbox}
\end{minipage}
\end{table}

\section{Comparison with State-Of-The-Art}\label{comparison}
We compare our methods on NTU-60 and NTU-120 w.r.t. other methods which make use of joint or joint+bones information on a one- or two-stream architecture, as we also did, for a fair comparison (Table \ref{table:3}). 
On NTU-60, ST-TR without bones outperforms all the state-of-the-art models not using bones, including 1s-AGCN and SAN \cite{san}, which uses self-attention too. Similarly, our ST-TR with bones outperforms all previous two-stream methods that use bones information as well, i.e., 2s-AGCN and 2s Shift-GCN. On NTU-120, the model based only on joint information outperforms all state-of-the-art methods making use of the same information. 
The competitive results validate the superiority of our method over architectures relying on convolution only.

\section{Conclusions}
In this paper, we propose a novel approach that introduces Transformer self-attention in skeleton activity recognition as an alternative to graph convolution. Through experiments on NTU-60 and NTU-120, we demonstrated that our Spatial Self-Attention module (SSA) can replace graph convolution, enabling more flexible and dynamic representations. Similarly, the Temporal Self-Attention module (TSA) overcomes the strict locality of standard convolution, leading to global motion pattern extraction. Moreover, our final Spatial-Temporal Transformer network (ST-TR) achieves state-of-the-art performance on NTU-RGB+D w.r.t. methods using same input information and streams setup.

%
%

\bibliographystyle{splncs04}
\bibliography{egbib}

\end{document}